\documentclass[11pt,a4paper]{article}
\pdfoutput=1
\usepackage[hyphens]{url}
\usepackage[hyperref]{acl2019}
\usepackage{times}
\usepackage{latexsym}

\usepackage{amsmath,amssymb}
\usepackage{multirow}
\usepackage{color}
\usepackage{graphicx}
\usepackage{bbm}
\usepackage{xspace}
\usepackage{wasysym}
\usepackage{algorithmic}
\usepackage{float}
\usepackage{mathtools}
\usepackage{pgfplots}
\pgfplotsset{compat=1.14}
\usepackage{placeins}
\usepackage{tabularx}
\usepackage{booktabs}
\usepgfplotslibrary{groupplots}
\usepackage{xcolor}
\usepackage{colortbl}
\usepackage[normalem]{ulem}

\newcommand{\source}{\mathbf{x}}

\newcommand{\goldreference}{\mathbf{t}}

\newcommand{\candhypos}{\mathcal{U}}
\newcommand{\candhypo}{\mathbf{u}}

\aclfinalcopy 
\def\aclpaperid{2575} 

\newcolumntype{L}[1]{>{\raggedright\arraybackslash}p{#1}}
\newcolumntype{C}[1]{>{\centering\arraybackslash}p{#1}}
\newcolumntype{R}[1]{>{\raggedleft\arraybackslash}p{#1}}\mathtoolsset{showonlyrefs}

\newenvironment{itemizesquish}{\begin{list}{\labelitemi}{\setlength{\itemsep}{-0.2em}\setlength{\labelwidth}{0.5em}\setlength{\leftmargin}{\labelwidth}
\addtolength{\leftmargin}{\labelsep}}}{\end{list}}

\newcommand{\lp}{\textsc{LP}\xspace}
\newcommand{\bp}{\textsc{BP}\xspace}
\newcommand{\simlength}{\textsc{SimiLe}\xspace}

\newcommand{\paragramphrase}{\textsc{paragram-phrase}\xspace}

\title{Beyond BLEU: \\ Training Neural Machine Translation with Semantic Similarity}

\author{John Wieting$^1$, Taylor Berg-Kirkpatrick$^2$, Kevin Gimpel$^3$, and Graham Neubig$^1$ \\
  $^1$Carnegie Mellon University,
  Pittsburgh, PA, 15213, USA \\
  $^2$University of California San Diego,
  San Diego, CA, 92093, USA\\
  $^3$Toyota Technological Institute at Chicago, Chicago, IL, 60637, USA \\
  {\small \texttt{\{jwieting,gneubig\}@cs.cmu.edu}, \texttt{tberg@eng.ucsd.edu},  \texttt{kgimpel@ttic.edu}}}
  
\date{}

\begin{document}
\maketitle
\begin{abstract}
While most neural machine translation (NMT) systems are still trained using maximum likelihood estimation, recent work has demonstrated that optimizing systems to directly improve evaluation metrics such as BLEU can substantially improve final translation accuracy. However, training with BLEU has some limitations: it doesn't assign partial credit, it has a limited range of output values, and it can penalize semantically correct hypotheses if they differ lexically from the reference.
In this paper, we introduce an alternative reward function for optimizing NMT systems that is based on recent work in semantic similarity.
We evaluate on four disparate languages translated to English, and find that training with our proposed metric results in 
better translations as evaluated by BLEU, semantic similarity, and human evaluation, and also that the optimization procedure converges faster.
Analysis suggests that this is because the proposed metric is more conducive to optimization, assigning partial credit and providing more diversity in scores than BLEU.%
\footnote{Code and data to replicate results are available at \url{https://www.cs.cmu.edu/~jwieting}.}
\end{abstract}

\section{Introduction}
In neural machine translation (NMT) and other natural language generation tasks, it is common practice to improve likelihood-trained models by further tuning their parameters to explicitly maximize an automatic metric of system accuracy -- for example, BLEU \cite{papineni2002bleu} or METEOR \cite{denkowski:lavie:meteor-wmt:2014}. Directly optimizing accuracy metrics involves backpropagating through discrete decoding decisions, and thus is typically accomplished with structured prediction techniques like reinforcement learning~\cite{Ranzato:ea:2016}, minimum risk training~\cite{shen2015minimum}, and other specialized methods \cite{wiseman2016beamsearchoptimization}.
Generally, these methods work by repeatedly generating a translation under the current parameters (via decoding, sampling, or loss-augmented decoding), comparing the generated translation to the reference, receiving some reward based on their similarity, and finally updating model parameters to increase future rewards.

In the vast majority of work, discriminative training has focused on optimizing BLEU (or its sentence-factored approximation). This is not surprising given that BLEU is the standard metric for system comparison at test time.
However, BLEU is not without problems when used as a training criterion. Specifically, since BLEU is based on n-gram precision, it aggressively penalizes lexical differences even when candidates might be synonymous with or similar to the reference: if an n-gram does not exactly match a sub-sequence of the reference, it receives no credit. While the pessimistic nature of BLEU differs from human judgments and is therefore problematic, it may, in practice, pose a more substantial problem for a different reason: BLEU is difficult to \textit{optimize} because it does not assign partial credit. As a result, learning cannot hill-climb through intermediate hypotheses with high synonymy or semantic similarity, but low n-gram overlap. Furthermore, where BLEU does assign credit, the objective is often flat: a wide variety of candidate translations can have the same degree of overlap with the reference and therefore receive the same score. This, again, makes optimization difficult because gradients in this region give poor guidance.

In this paper we propose \simlength, a simple alternative to matching-based metrics like BLEU for use in discriminative NMT training. As a new reward, we introduce a measure of \emph{semantic similarity between the generated hypotheses and the reference translations} evaluated by an embedding model trained on a large external corpus of paraphrase data. Using an embedding model to evaluate similarity allows the range of possible scores to be continuous and, as a result, introduces fine-grained distinctions between similar translations. This allows for partial credit and reduces the penalties on semantically correct but lexically different translations. Moreover, since the output of \simlength is continuous, it provides more informative gradients during the optimization process by distinguishing between candidates that would be similarly scored under matching-based metrics like BLEU. Lastly, we show in our analysis that \simlength has an additional benefit over BLEU by translating words with heavier semantic content more accurately.

To define an exact metric, we reference the burgeoning field of research aimed at measuring semantic textual similarity (STS) between two sentences~\cite{le2014distributed,pham-EtAl:2015:ACL-IJCNLP,wieting-16-full,hill2016learning,conneau2017supervised,pagliardini2017unsupervised}. Specifically, we start with the method of \newcite{wieting2017pushing}, which learns paraphrastic sentence representations using a contrastive loss and a parallel corpus induced by backtranslating bitext. Wieting and Gimpel showed that simple models that average word or character trigram embeddings
can be highly effective for semantic similarity. The strong performance,
domain robustness, and computationally efficiency of these models make them good candidates for experimenting with incorporating semantic similarity into neural machine translation. For the purpose of discriminative NMT training, we augment these basic models with two modifications: we add a length penalty to avoid short translations, and calculate similarity by composing the embeddings of subword units, rather than words or character trigrams. We find that using subword units also yields better performance on the STS evaluations and is more efficient than character trigrams.

We conduct experiments with our new metric on the 2018 WMT~\cite{WMT2018Findings} test sets, translating four languages, Czech, German, Russian, and Turkish, into English. Results demonstrate that optimizing \simlength during training results in not only improvements in the same metric during test, but also in consistent improvements in BLEU. Further, we conduct a human study to evaluate system outputs and find significant improvements in human-judged translation quality for all but one language. Finally, we provide an analysis of our results in order to give insight into the observed gains in performance. Tuning for metrics other than BLEU has not (to our knowledge) been extensively examined for NMT, and we hope this paper provides a first step towards broader consideration of training metrics for NMT.

\section{\simlength Reward Function} \label{sec:models}
Since our goal is to develop a continuous metric of sentence similarity, we borrow from a line of work focused on domain agnostic semantic similarity metrics.  
We motivate our choice for applying this line of work to training translation models in Section 2.1. Then in Section 2.2, we describe how we train our similarity metric (SIM), how we compute our length penalty, and how we tie these two terms together to form \simlength.

\subsection{\simlength}

Our \simlength
metric is based on the sentence similarity metric of \citet{wieting2017pushing}, which we choose as a starting point because it has state-of-the-art unsupervised performance on a host of domains for semantic textual similarity.\footnote{In semantic textual similarity the goal is to 
produce scores that correlate with human judgments on the degree to which two sentences have the same semantics. In embedding based models, including the models used in this paper, the score is produced by the cosine of the two sentence embeddings.} Being both unsupervised and domain agnostic provide evidence that the model generalizes well to unseen examples. This is in contrast to supervised methods which are often imbued with the bias of their training data.

\paragraph{Model.}
Our sentence encoder $g$ 
averages 300 dimensional subword unit\footnote{We use \texttt{sentencepiece} which is available at \url{https://github.com/google/sentencepiece}. We limited the vocabulary to 30,000 tokens.} embeddings to create a sentence representation. 
The similarity of two sentences, SIM, is obtained by encoding both with $g$ and then calculating their cosine similarity.

\paragraph{Training.} We follow \citet{wieting2017pushing} in learning the parameters of the encoder 
$g$. The training data is a set $S$ of paraphrase pairs\footnote{We use 16.77 million paraphrase pairs filtered from the ParaNMT corpus~\cite{wieting2017pushing}. The corpus is filtered by a sentence similarity score based on the \paragramphrase from \citet{wieting-16-full} and word trigrams overlap, which is calculated by counting word trigrams in the reference and translation, then dividing the number of shared trigrams by the total number in the reference or translation, whichever has fewer. These form a balance between semantic similarity (similarity score) and diversity (trigram overlap). We kept all sentences in ParaNMT with a similarity score $\geq$ 0.5 and a trigram overlap score $\leq$ 0.2. Recently, in~\cite{wieting2019simple} it has been shown that strong performance on semantic similarity tasks can also be achieved using bitext directly without the need for backtranslation.} $\langle s, s'\rangle$ and we use a margin-based loss:
\begin{align*}
\ell(s, s') = 
&\max(0,\delta - \cos(g(s), g(s'))\\
& + \cos(g(s), g(t)))
\label{eq:obj}
\end{align*} 
\noindent where 
$\delta$ is the margin, 
and
$t$ is a {\it negative example}. 
The intuition is that we want the two texts to be more similar to each other than to their negative examples.
To select $t$,
we choose the most similar sentence in a collection of mini-batches called a {\it mega-batch}. 

\begin{table}
\setlength{\tabcolsep}{5pt}
\scriptsize
\centering
\begin{tabular} { |l|l| l | c | c | c | c | c |}
\hline
Model & 2012 & 2013 & 2014 & 2015 & 2016  \\
\hline
SIM (300 dim.) & 69.2 & 60.7 & 77.0 & 80.1 & \bf 78.4 \\
\simlength & \bf 70.1 & 59.8 & 74.7 & 79.4 & 77.8 \\
\hline
\citet{wieting2017pushing} & 67.8 & \bf 62.7 & \bf 77.4 & \bf 80.3 & 78.1 \\
\hline
BLEU & 58.4 & 37.8 & 55.1 & 67.4 & 61.0 \\
BLEU (symmetric) & 58.2 & 39.1 & 56.2 & 67.8 & 61.2 \\
METEOR & 53.4 & 47.6 & 63.7 & 68.8 & 61.8 \\
METEOR (symmetric) & 53.8 & 48.2 & 65.1 & 70.0 & 62.7 \\
\hline
STS $1^{\mathit{st}}$ Place & 64.8 & 62.0 & 74.3 & 79.0 & 77.7 \\
STS $2^{\mathit{nd}}$ Place & 63.4 & 59.1 & 74.2 & 78.0 & 75.7 \\
STS $3^{\mathit{rd}}$ Place & 64.1 & 58.3 & 74.3 & 77.8 & 75.7 \\
\hline
\end{tabular}
\caption{\label{table:sim} Comparison of the semantic similarity model used in this paper (SIM) with a number of strong baselines including the model of~\cite{wieting2017pushing} and the top 3 performing STS systems for each year. Symmetric refers to taking the average score of the metric with each sentence having a turn in the reference position.
}
\end{table}

\begin{table}
\setlength{\tabcolsep}{5pt}
\small
\centering
\begin{tabular} { |l|c|c|}
\hline
Model & newstest2015 & newstest2016 \\
\hline
SIM & 58.2 & 53.1 \\
\simlength & 58.4 & 53.2 \\
BLEU & 53.6 & 50.0 \\
METEOR & \bf 58.9 & \bf 57.2 \\
\hline
\end{tabular}
\caption{\label{table:mteval} Comparison of models on machine translation quality evaluation datasets. Scores are in Spearman's $\rho$.
}
\end{table}

Finally, we note that SIM is robust to domain, as shown  
by its strong performance on the STS tasks which cover a broad range of domains. 
We note that SIM was trained primarily on subtitles, while we use news data to train and evaluate our NMT models. Despite this domain switch, we are able to show improved performance over a baseline using BLEU, providing more evidence of the robustness of this method.

\paragraph{Length Penalty.}

Our initial experiments showed that when using just the similarity metric, SIM, there was nothing preventing the model from learning to generate long sentences, often at the expense of repeating words. This is the opposite case from BLEU, where the n-gram precision is not penalized for generating too few words. Therefore, in BLEU, a brevity penalty (BP) was introduced to penalize sentences when they are shorter than the reference. The penalty is:
$$
\bp(r,h) = e^{1-\frac{|r|}{|h|}}
$$
where $r$ is the reference and $h$ is the generated hypothesis, with $|r|$ and $|h|$ their respective lengths. We experimented with modifying this penalty to only penalize generated sentences that are longer than the target (so we switch $r$ and $h$ in the equation). However, we found that this favored short sentences. We instead penalize a generated sentence if its length differs at all from that of the target. Therefore, our length penalty is:
$$
\lp(r,h) = e^{1-\frac{\max(|r|,|h|)}{\min(|r|,|h|)}}
$$
\paragraph{{\simlength}.}
Our final metric, which we refer to as \simlength, is defined as follows:
\begin{equation}
\text{\simlength} = \text{\lp}(r,h)^{\alpha}\text{SIM}(r,h)
\end{equation}
In initial experiments we found that performance could be improved slightly by lessening the influence of \lp, so we  
fix $\alpha$ to be 0.25.

\subsection{Motivation}

There is a vast literature on metrics for \emph{evaluating} machine translation outputs automatically (For instance, WMT metrics task papers like~\citet{bojar2017findings}). In this paper we demonstrate that \emph{training} towards metrics other than BLEU has significant practical advantages in the context of NMT. While this could be done with any number of metrics, in this paper we experiment with a single semantic similarity metric, and due to resource constraints leave a more extensive empirical comparison of other evaluation metrics to future work. That said, we designed \simlength as a semantic similarity model with high accuracy, domain robustness, and computational efficiency to be used in minimum risk training for machine translation.\footnote{\simlength, including time to segment the sentence, is about 20 times faster than METEOR when code is executed on a GPU (NVIDIA GeForce GTX 1080).}

While semantic similarity is not an exact replacement for measuring machine translation quality, we argue that it serves as a decent proxy at least as far as minimum risk training is concerned. To test this, we compare the similarity metric term in \simlength (SIM) to BLEU and METEOR on two machine quality datasets\footnote{We used the segment level data from newstest2015 and newstest2016 available at \url{http://statmt.org/wmt18/metrics-task.html}. The former contains 7 language pairs and the latter 5.} and report their correlation with human judgments in Table~\ref{table:mteval}. Machine translation quality measures account for more than semantics as they also capture other factors like fluency.  A manual error analysis and the fact that the machine translation correlations in Table~\ref{table:mteval} are close, but the semantic similarity correlations\footnote{Evaluation is on the SemEval Semantic Textual Similarity (STS) datasets from 2012-2016~\cite{agirre2012semeval,agirre2013sem,agirre2014semeval,agirre2015semeval,agirre2016semeval}. In the SemEval STS competitions, teams 
create models that need to work well 
on domains both represented in the training data and hidden domains revealed at test time. Our model and those of \citet{wieting2017pushing}, in contrast to the best performing STS systems, do not use any 
manually-labeled training examples nor any other linguistic resources beyond the ParaNMT corpus~\cite{wieting2017pushing}.} in Table~\ref{table:sim} are not, suggest that the difference between METEOR and SIM largely lies in fluency. 
However, not capturing fluency is something that can be ameliorated 
by adding a down-weighted maximum-likelihood (MLE) loss to the minimum risk loss. This  was done by  \citet{edunov2018classical}, and we use this in our experiments as well. 

\section{Machine Translation Preliminaries} \label{sec:translation}
\paragraph{Architecture.}

Our model and optimization procedure are based on prior work on structured prediction training for neural machine translation~\cite{edunov2018classical} and are implemented in Fairseq.\footnote{\url{https://github.com/pytorch/fairseq}}
Our architecture follows the paradigm of an encoder-decoder with soft attention~\cite{bahdanau2014neural} and we use the same architecture for each language pair in our experiments. We use gated convolutional encoders and decoders~\cite{gehring2017convolutional}. We use 4 layers for the encoder and 3 for the decoder, setting the hidden state size for all layers to 256, and the filter width of the kernels to 3. We use byte pair encoding~\cite{sennrich2015neural-arxiv}, with a vocabulary size of 40,000 for the combined source and target vocabulary. The dimension of the BPE embeddings is set to 256.

\paragraph{Objective Functions.}

Following~\cite{edunov2018classical}, we first train models with maximum-likelihood with label-smoothing ($\mathcal{L}_{\text{TokLS}}$)~\cite{szegedy2016rethinking,pereyra2017regularizing}. We set the confidence penalty of label smoothing to be 0.1. Next, we fine-tune the model with a weighted average of minimum risk training ($\mathcal{L}_{\text{Risk}}$)~\cite{shen2015minimum} and ($\mathcal{L}_{\text{TokLS}}$), where the expected risk is defined as:
\begin{equation}
\mathcal{L}_{\text{Risk}} =
\sum_{\candhypo \in \candhypos(\source)} \operatorname{cost}(\goldreference, \candhypo) \frac{p(\candhypo|\source)}{\sum_{\candhypo' \in \candhypos(\source)} p(\candhypo'|\source)}
\end{equation}
where $\candhypo$ is a candidate hypothesis, $\candhypos(\source)$ is a set of candidate hypotheses, and $\goldreference$ is the reference. Therefore, our fine-tuning objective becomes:
$$
\mathcal{L}_{\text{Weighted}} = \gamma \mathcal{L}_{\text{TokLS}} + (1-\gamma)\mathcal{L}_{\text{Risk}}
$$
\noindent We tune $\gamma$ from the set $\{0.2,0.3,0.4\}$ in our experiments. In minimum risk training, we aim to minimize the expected cost. In our case that is $1-\text{BLEU}(t,h)$ or $1-\text{\simlength}(t,h)$ where $t$ is the target and $h$ is the generated hypothesis. As is commonly done, we use a smoothed version of BLEU by adding 1 to all $n$-gram counts except unigram counts. This is to prevent BLEU scores from being overly sparse \cite{lin04orange}.
We generate candidates for minimum risk training from $n$-best lists with 8 hypotheses and do not include the reference in the set of candidates. 

\paragraph{Optimization.}
We optimize our models using Nesterov's accelerated gradient method~\cite{sutskever2013importance} using a learning rate of 0.25 and momentum of 0.99. Gradients are renormalized to norm 0.1~\cite{pascanu2012difficulty}. We train the $\mathcal{L}_{\text{TokLS}}$ objective for 200 epochs and the combined objective, $\mathcal{L}_{\text{Weighted}}$, for 10. Then for both objectives, we anneal the learning rate by reducing it by a factor of 10 after each
epoch until it falls below $10^{-4}$. Model selection is done by selecting the model with the lowest validation loss on the validation set. To select models across the different hyperparameter settings, we chose the model with the highest performance on the validation set for the evaluation being considered.

\section{Experiments}  \label{sec:exp}
\subsection{Data}

\begin{table}
\centering
\small
\begin{tabular} { | l | c | c | c | }
\hline
Lang. & Train & Valid & Test \\
\hline
\texttt{cs-en} & 218,384 & 6,004 & 2,983\\
\texttt{de-en} & 284,286 & 7,147 & 2,998\\
\texttt{ru-en} & 235,159 & 7,231 & 3,000\\
\texttt{tr-en} & 207,678 & 7,008 & 3,000\\
\hline
\end{tabular}
\caption{\label{table:data} Number of sentence pairs in the training/validation/test sets for all four languages.
}
\end{table}

\begin{table*}[th!]
\centering
\small
\begin{tabular} { | l | c | c | c | c | c | c | c | c |}
\hline
& \multicolumn{2}{c|}{\texttt{de-en}} & \multicolumn{2}{c|}{\texttt{cs-en}} & \multicolumn{2}{c|}{\texttt{ru-en}} & \multicolumn{2}{c|}{\texttt{tr-en}}\\
\hline
Model & BLEU & SIM & BLEU & SIM & BLEU & SIM & BLEU & SIM \\
\hline
MLE & 27.52\phantom{*}\phantom{*} & 76.19\phantom{*}\phantom{*} & 17.02\phantom{*}\phantom{*} & 67.55\phantom{*}\phantom{*} & 17.92\phantom{*}\phantom{*} & 69.13\phantom{*}\phantom{*} & 14.47\phantom{*}\phantom{*} & 65.97\phantom{*}\phantom{*}\\
BLEU & 27.92$^{\ddagger}$\phantom{*} & 76.28$^{\ddagger}$\phantom{*} & 17.38$^{\ddagger}$\phantom{*} & 67.87$^{\ddagger}$\phantom{*} & 17.97\phantom{*}\phantom{*} & 69.29$^{\ddagger}$\phantom{*} & 15.10$^{\ddagger}$\phantom{*} & 66.53$^{\ddagger}$\phantom{*} \\
\simlength & \bf 28.56$^{\dagger\ddagger}$ & \bf 77.52$^{\dagger\ddagger}$ & \bf 17.60$^{\dagger\ddagger}$ & \bf 68.89$^{\dagger\ddagger}$ & \bf 18.44$^{\dagger\ddagger}$ & \bf 70.69$^{\dagger\ddagger}$ & \bf 15.47$^{\dagger\ddagger}$ & \bf 67.76$^{\dagger\ddagger}$ \\
Half & 28.25$^{\dagger\ddagger}$ & 76.92$^{\dagger\ddagger}$ & 17.52$^{\dagger\ddagger}$ & 68.26$^{\dagger\ddagger}$ & 18.26$^{\dagger\ddagger}$ & 70.32$^{\dagger\ddagger}$ & 15.40$^{\dagger\ddagger}$ & 67.14$^{\dagger\ddagger}$ \\
\hline
\end{tabular}
\caption{\label{table:bleusimresults} Results on translating four languages to English for MLE, BLEU, \simlength and Half.
$\dagger$ denotes statistical significance ($p < 0.05$) over BLEU and $\ddagger$ denotes statistical significance over MLE. Statistical significance was computed using paired bootstrap resampling~\cite{koehn2004statistical}.}
\end{table*}

Training models with minimum risk is expensive, but we wanted to evaluate in a difficult, realistic setting using a diverse set of languages. Therefore, we experiment on four language pairs: Czech (\texttt{cs-en}), German (\texttt{de-en}), Russian (\texttt{ru-en}), and Turkish (\texttt{tr-en}) translating to English (\texttt{en}). For training data, we use News Commentary v13\footnote{\url{http://data.statmt.org/wmt18/translation-task/training-parallel-nc-v13.tgz}} provided by WMT~\cite{WMT2018Findings} for \texttt{cs-en}, \texttt{de-en}, and \texttt{ru-en}. For training the Turkish system, we used the WMT 2018 parallel data which consisted of the SETIMES2\footnote{\url{http://opus.lingfil.uu.se/SETIMES2.php}} corpus. The validation and development sets for \texttt{de-en}, \texttt{cs-en}, and \texttt{ru-en} were the WMT 2016 and WMT 2017 validation sets. For \texttt{tr-en}, the validation set was the WMT 2016 validation set and the WMT 2017 validation and test sets. Test sets for each language were the official WMT 2018 test sets.

\subsection{Automatic Evaluation} 

We first use corpus-level BLEU and the corpus average SIM score to evaluate the outputs of the different experiments. It is important to note that in this case, SIM is not the same as \simlength.  SIM is only the semantic similarity component of \simlength and therefore lacks the length penalization term. We used this metric to estimate the degree to which the semantic content of a translation and its reference overlap. When evaluating semantic similarity, we find that SIM outperforms \simlength marginally as shown in Table~\ref{table:sim}.

We compare systems trained with 4 objectives: 
\begin{itemizesquish}
\item MLE: Maximum likelihood with label smoothing
\item BLEU: Minimum risk training with 1-BLEU as the cost
\item \simlength: Minimum risk training with 1-\simlength as the cost
\item Half: Minimum risk training with a new cost that is half BLEU and half \simlength:  $1-\frac{1}{2}(\text{BLEU} + \simlength)$
\end{itemizesquish}

The results are shown in Table~\ref{table:bleusimresults}. From the table, we see that using \simlength performs the best when using BLEU and SIM as evaluation metrics for all four languages. It is interesting that using \simlength in the cost leads to larger BLEU improvements than using BLEU alone, the reasons for which we examine further in the following sections. It is important to emphasize that increasing BLEU was not the goal of our proposed method, human evaluations were our target, but this is a welcome surprise. Similarly, using BLEU as the cost function leads to large gains in SIM, though these gains are not as large as when using \simlength in training. 

\subsection{Human Evaluation}

We also perform human evaluation, comparing MLE training with minimum risk training using \simlength and BLEU as costs. We selected 200 sentences along with their translation from the respective test sets of each language. The sentences were selected nearly randomly with the only constraints that they be between 3 and 25 tokens long and also that the outputs for \simlength and BLEU were not identical. The translators then assigned a score from 0-5 based on how well the translation conveyed the information contained in the reference.\footnote{Wording of the evaluation is available in Section~\ref{appendix}.
}

\begin{table}
\centering
\small
\begin{tabular} { | l | c | c | c |}
\hline
& \multicolumn{3}{c|}{Avg. Score} \\
\hline
Lang. & MLE & BLEU & \simlength \\
\hline
 \texttt{cs-en}  & 0.98\phantom{*} & 0.90\phantom{*} & \bf 1.02$^\dagger$\phantom{*}\\
 \texttt{de-en}  & 0.93\phantom{*} & 0.85\phantom{*} & \bf 1.00$^{\dagger}$\phantom{*}\\
 \texttt{ru-en} & 1.22\phantom{*} & 1.21\phantom{*} & \bf 1.31$^{\dagger\ddagger}$\\
\texttt{tr-en} & 0.98$^*$ & \bf 1.03$^*$ & 0.78\phantom{*}\phantom{*}\\
\hline
\end{tabular}
\caption{\label{table:human} Average human ratings on 200 sentences from the test set for each of the respective languages. $\dagger$ denotes statistical significance ($p < 0.05$) over BLEU, except for the case of \texttt{cs-en}, where $p=0.06$. $\ddagger$ denotes statistical significance over MLE, and * denotes statistical significance over \simlength. Statistical significance was computed using paired bootstrap resampling.}
\end{table}

From the table, we see that minimum risk training with \simlength as the cost scores the highest across all language pairs except Turkish. Turkish is also the language with the lowest test BLEU (See Table~\ref{table:bleusimresults}). An examination of the human-annotated outputs shows that in Turkish (unlike the other languages) repetition was a significant problem for the \simlength system in contrast to MLE or BLEU. We hypothesize that one weakness of \simlength may be that it needs to start with some minimum level of translation quality in order to be most effective. The biggest improvement over BLEU is on \texttt{de-en} and \texttt{ru-en}, which have the highest MLE BLEU scores in Table~\ref{table:bleusimresults} which further lends credence to this hypothesis.

\section{Quantitative Analysis}
We next analyze our model using the validation set of the \texttt{de-en} data unless stated otherwise. We chose this dataset for the analysis since it had the highest MLE BLEU scores of the languages studied.

\subsection{Partial Credit}

\begin{figure}
    \centering
    \small
    \includegraphics[width=0.5\textwidth]{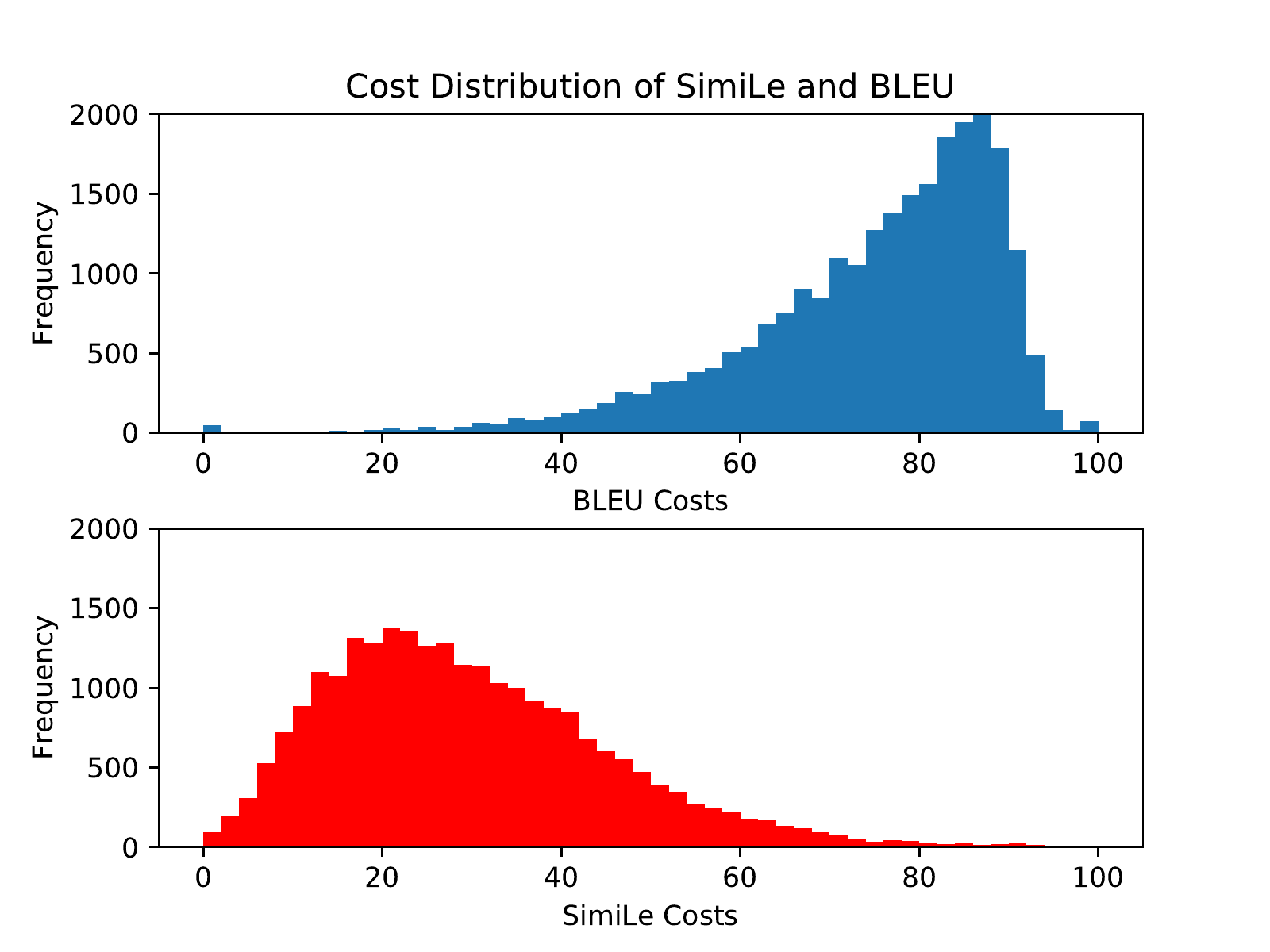}
    \caption{Distribution of scores for \simlength and BLEU.}
    \label{fig:scores}
\end{figure}

We analyzed the distribution of the cost function for both \simlength and BLEU on the \texttt{de-en} validation set before any fine-tuning. Again, using an $n$-best list size of 8, we computed the cost for all generated translations and plotted their histogram in Figure~\ref{fig:scores}. 
The plots show that the distribution of scores for \simlength and BLEU are quite different. Both distributions are not symmetrical Gaussian, however the distribution of BLEU scores is significantly more skewed with much higher costs. This tight clustering of costs provides less information during training.

Next, for all $n$-best lists, we computed all differences between scores of the hypotheses in the beam. Therefore, for a beam size of 8, this results in 28 different scores. We found that of the 86,268 scores, the difference between scores in an $n$-best list is $\geq 0$ 99.0\% of the time for \simlength, but 85.1\% of the time for BLEU. The average difference is 4.3 for BLEU and 4.8 for \simlength, showing that \simlength makes finer grained distinctions among candidates.

\begin{figure}[h]
    \centering
    \small
    \includegraphics[width=0.5\textwidth]{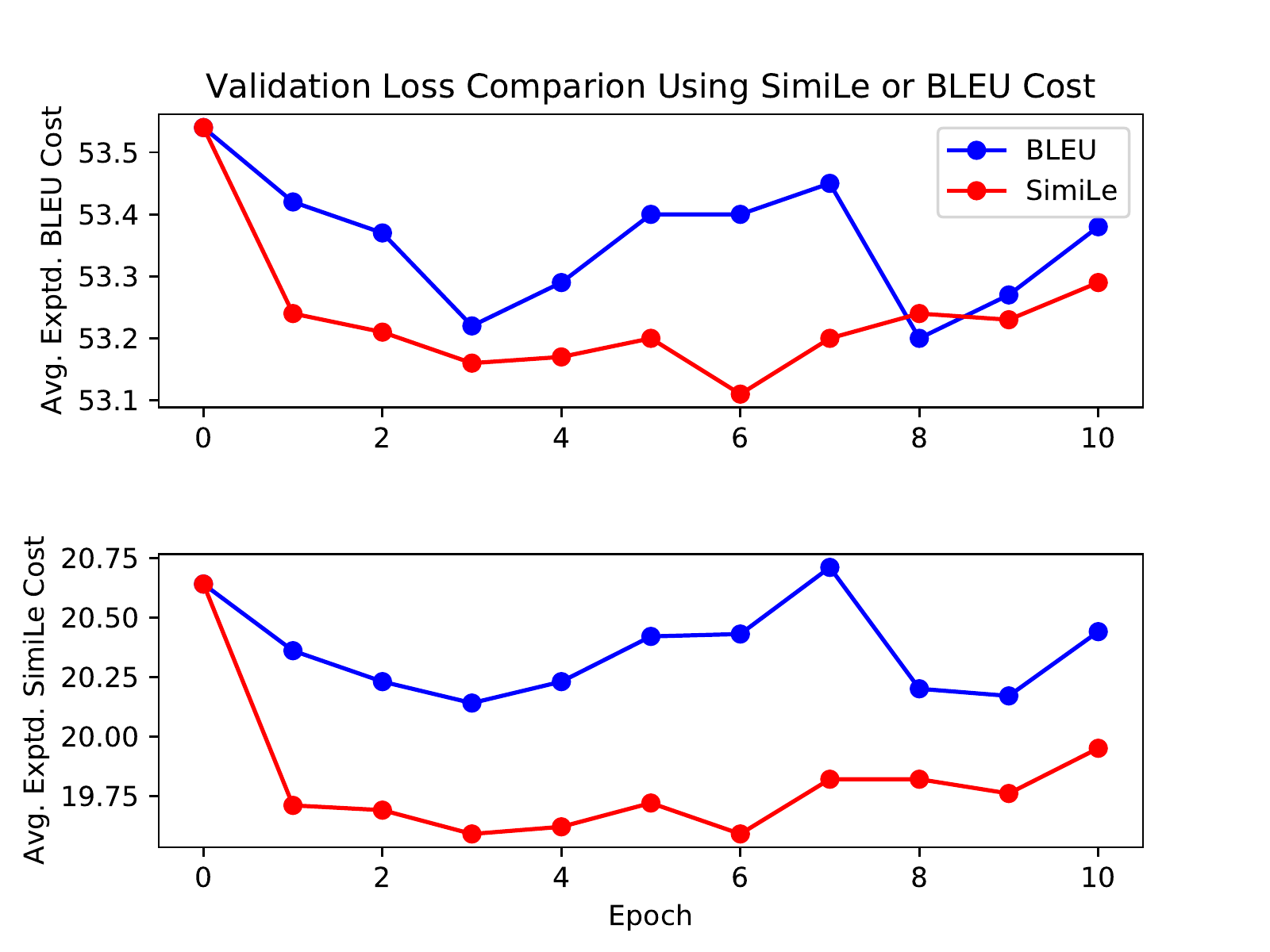}
    \caption{Validation loss comparison for \simlength and BLEU. The top plot shows the expected BLEU cost when training with BLEU and \simlength. The bottom plot shows the expected \simlength cost when training with BLEU and \simlength.}
    \label{fig:validplot}
\end{figure}

\subsection{Validation Loss}

We next analyze the validation loss during training of the \texttt{de-en} model for both using \simlength and BLEU as costs. We use the hyperparameters of the model with the highest BLEU on the validation set for model selection. Since the distributions of costs vary significantly between \simlength and BLEU, with BLEU having much higher costs on average, we compute the validation loss with respect to both cost functions for each of the two models. 

In Figure~\ref{fig:validplot}, we plot the risk objective 
for the first 10 epochs of training. In the top plot, we see that the risk objective for both BLEU and \simlength decreases much faster when using \simlength to train than BLEU. The expected BLEU also reaches a significantly lower value on the validation set when training with \simlength. The same trend occurs in the lower plot, this time measuring the expected \simlength cost on the validation set. 

From these plots, we see that optimizing with \simlength results in much faster training. It also reaches a lower validation loss, and from Table~\ref{table:bleusimresults}, we've already shown that the \simlength and BLEU on the test set are higher for models trained with \simlength. To hammer home the point at how much faster the models trained with \simlength reach better performance, we evaluated after just 1 epoch of training and found that the model trained with BLEU had SIM/BLEU scores of 86.71/27.63 while the model trained with \simlength had scores of 87.14/28.10. A similar trend was observed in the other language pairs as well, where the validation curves show a much larger drop-off after a single epoch when training with \simlength than with BLEU.

\subsection{Effect of $n$-best List Size}

\begin{figure}[]
    \centering
    \small
    \includegraphics[width=0.5\textwidth]{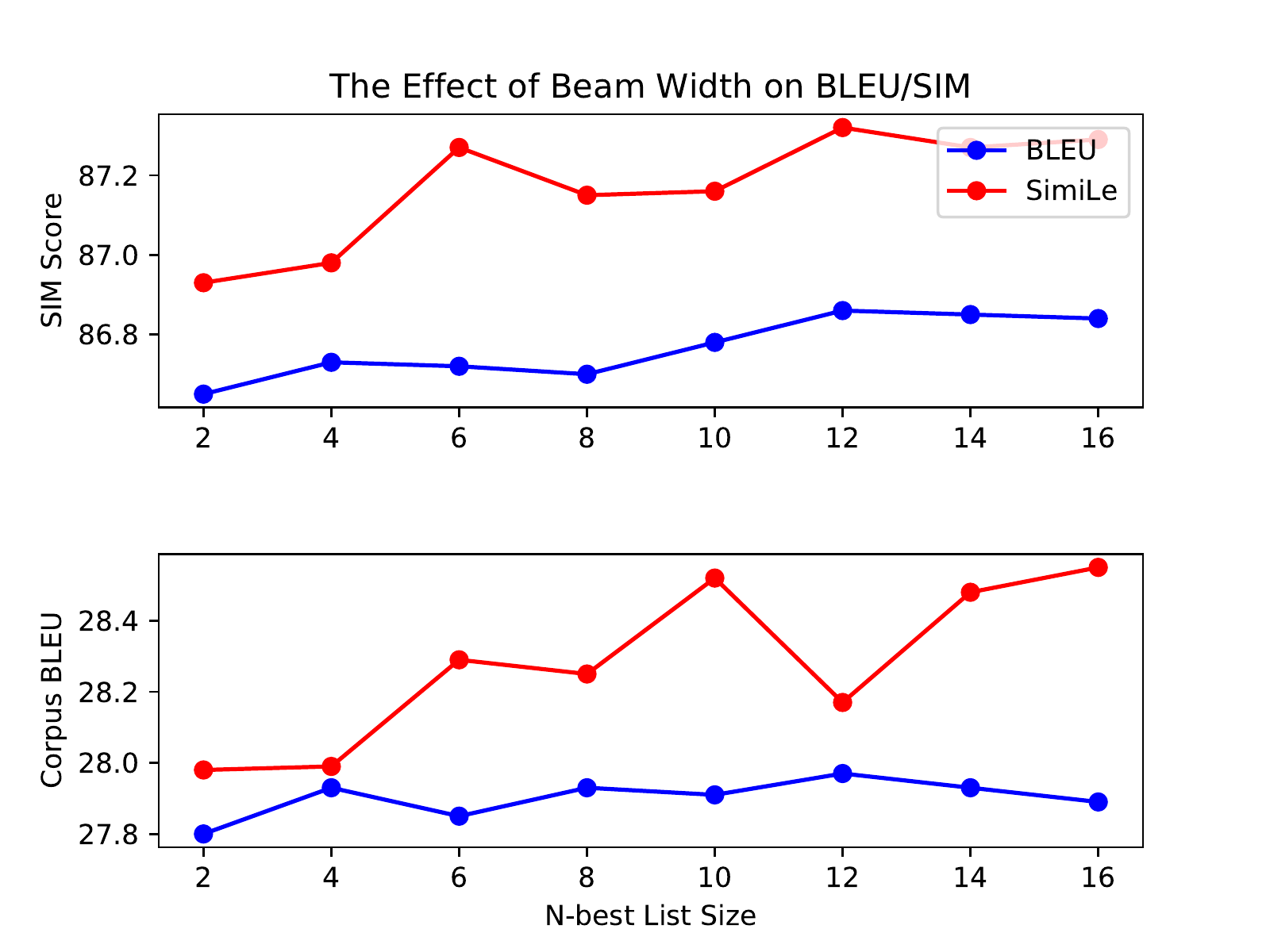}
    \caption{The relationship between $n$-best list size and performance as measured by average SIM score or corpus-level BLEU when training using \simlength or BLEU as a cost.}
    \label{fig:beamplot}
\end{figure}

\renewcommand{\tabcolsep}{4pt}
\begin{table}
\centering
\scriptsize
\begin{tabular} { | l | l | l | l | l | l |}
\hline
Lang./Bucket & \texttt{cs-en} $\Delta$ & \texttt{de-en} $\Delta$ & \texttt{ru-en} $\Delta$ & \texttt{tr-en} $\Delta$ & Avg. \\
\hline
1 & \phantom{-}0.1 & \phantom{-}0.8 & \phantom{-}0.2 & \phantom{-}0.1 & 0.30\\
2-5 & \phantom{-}1.2 & \phantom{-}0.6 & \it \phantom{-}0.0 & \phantom{-}0.2 & 0.50\\
6-10 & \phantom{-}0.4 &\phantom{-}0.7 & \phantom{-}1.4 & \it -0.3 & 0.55\\
11-100 & \phantom{-}0.2 & \phantom{-}0.6 & \phantom{-}0.6 & \phantom{-}0.4 & 0.45\\
101-1000 & \it -0.3 & \phantom{-}0.3 & \phantom{-}0.4 & \phantom{-}0.2 & 0.15\\
1001+ & \it -0.2 & \phantom{-}0.5 & \phantom{-}0.4 & \it -0.0 & 0.08\\
\hline
DET & \phantom{-}0.1 & \it -0.1 & \phantom{-}0.7 & -0.5 & 0.03\\
PRON & \phantom{-}0.6 & \it -0.3 & \phantom{-}0.1 & \phantom{-}0.9 & 0.33\\
PREP & \phantom{-}0.2 & \it -0.3 & \phantom{-}0.5 & \phantom{-}0.5 & 0.24\\
CONJ & \phantom{-}0.1 & \phantom{-}1.1 & \phantom{-}0.3 & -0.5 & 0.27\\
PUNCT & \it -0.4 & \phantom{-}1.3 & \phantom{-}0.8 & -0.4 & 0.34\\
NUM & \phantom{-}0.6 & \phantom{-}2.2 & \phantom{-}1.8 & \phantom{-}1.3 & \bf 1.48\\
SYM & \phantom{-}0.3 & \phantom{-}3.6 & \phantom{-}4.4 & \phantom{-}1.7 & \bf 2.50\\
INTJ & \phantom{-}3.2 & \it -1.1 & \phantom{-}3.2 & \it -2.6 & \bf 0.66\\
VERB & \phantom{-}0.2 & \phantom{-}0.3 & \phantom{-}0.0 & \it \phantom{-}0.0 & 0.13\\
ADJ & \phantom{-}0.2 & \phantom{-}0.7 & \phantom{-}0.3 & \it -0.2 & 0.25\\
ADV & \it -0.2 & \phantom{-}0.1 & \phantom{-}0.8 & \phantom{-}0.7 & 0.34\\
NOUN & \phantom{-}0.3 & \phantom{-}1.1 & \phantom{-}0.8 & \phantom{-}0.4 & \bf 0.63\\
PRNOUN & \phantom{-}0.5 & \phantom{-}1.2 & \phantom{-}0.6 & \phantom{-}0.4 & \bf 0.65\\
\hline
\end{tabular}
\caption{\label{table:lexical} Difference in F1 score for various buckets of words. The values in the table are the difference between the F1 obtained when training using \simlength and when training using BLEU (positive values means \simlength had a higher F1). The first part of the table shows F1 scores across bins defined by word frequency on the test set. So words appearing only 1 time are in the first row, between 2-5 times are in the second row, etc. The next part of the table buckets words by coarse part-of-speech tags.}
\end{table}

As mentioned in Section~\ref{sec:translation}, we used an $n$-best list size of 8 in our minimum risk training experiments. In this section, we train \texttt{de-en} translation models with various $n$-best list sizes and investigate the relationship between beam size and test set performance when using \simlength or BLEU as a cost. We hypothesize that since BLEU is not as fine-grained a metric as \simlength, expanding the number of candidates would close the gap between BLEU and \simlength as BLEU would have access to a more candidates with more diverse scores.
The results of our experiment on the are shown in Figure~\ref{fig:beamplot} and show that models trained with \simlength actually improve in BLEU and SIM more significantly as $n$-best list size increases. This is possibly due to small $n$-best sizes inherently upper-bounding performance regardless of training metric, and \simlength being a better measure overall when the $n$-best is sufficiently large to learn. 

\begin{table*}
    \centering
    \scriptsize
    \begin{tabular}{|p{3.5cm}p{1cm}C{1.4cm}p{7.5cm}|}
    \hline
        Reference & System & Human Score & Translation \\
    \hline
        \multirow{3}{*}{\parbox{3.5cm}{I will tell you my personal opinion of him.}}. & BLEU & 2 & I will have a personal opinion on it. \\
        & \simlength & 4 & I will tell my personal opinion about it. \\
        & MLE & 2 & I will have a personal view of it. \\
    \hline
        \multirow{3}{*}{\parbox{3.5cm}{In my case, it was very varied.}} & BLEU & 0 & I was very different from me.\\
        & \simlength & 4 & For me, it was very different.\\
        & MLE & 1 & In me, it was very different.\\
    \hline
        \multirow{3}{*}{\parbox{3.5cm}{We're making the city liveable.}} & BLEU & 0 & We make the City of Life Life.\\
        & \simlength & 3 & We make the city viable.\\
        & MLE & 0 & We make the City of Life.\\
    \hline
        \multirow{3}{*}{\parbox{3.5cm}{The head of the White House said that the conversation was ridiculous.}} & BLEU & 0 & The White House chairman, the White House chip called a ridiculous.\\
        & \simlength & 4 & The White House's head, he described the conversation as ridiculous.\\
        & MLE & 1 & The White House chief, he called the White House, he called a ridiculous.\\
    \hline
        \multirow{3}{*}{\parbox{3.5cm}{According to the former party leaders, so far the discussion “has been predominated by expressions of opinion based on emotions, without concrete arguments”.}} & BLEU & 3 & According to former party leaders, the debate has so far had to be "elevated to an expression of opinion without concrete arguments."\\
        & \simlength & 5 & In the view of former party leaders, the debate has been based on emotions without specific arguments."\\
        & MLE & 4 & In the view of former party leaders, in the debate, has been based on emotions without specific arguments."\\
    \hline
        \multirow{3}{*}{\parbox{3.5cm}{We are talking about the 21st century: servants.}} & BLEU & 4 & We are talking about the 21st century: servants.\\
        & \simlength & 1 & In the 21st century, the 21st century is servants.\\
        & MLE & 0 & In the 21st century, the 21st century is servants.\\
    \hline
        \multirow{3}{*}{\parbox{3.5cm}{Prof. Dr. Caglar continued:}} & BLEU & 3 & They also reminded them.\\
        & \simlength & 0 & There are no Dr. Caglar.\\
        & MLE & 3 & They also reminded them.\\
    \hline
    \end{tabular}
    \caption{Translation examples for min-risk models trained with \simlength and BLEU and our baseline MLE model. 
    }
    \label{table:examples}
\end{table*}

\subsection{Lexical F1}

We next attempt to elucidate exactly which parts of the translations are improving due to using \simlength cost compared to using BLEU. We compute the F1 scores for target word types based on their frequency and their coarse part-of-speech tag (as labeled by SpaCy\footnote{ \url{https://github.com/explosion/spaCy}}) on the test sets for each language and show the results in Table~\ref{table:lexical}.\footnote{We use \texttt{compare-mt} \cite{neubig19naacl} available at \url{https://github.com/neulab/compare-mt}.}

From the table, we see that training with \simlength helps produce low frequency words more accurately, a fact that is consistent with the part-of-speech tag analysis in the second part of the table. \citet{wieting-17-full} noted that highly discriminative parts-of-speech, such as nouns, proper nouns, and numbers, made the most contribution to the sentence embeddings. Other works~\cite{pham-EtAl:2015:ACL-IJCNLP,wieting-16-full} have also found that when training semantic embeddings using an averaging function, embeddings that bear the most information regarding the meaning have larger norms. We also see that these same parts-of-speech (nouns, proper nouns, numbers) have the largest difference in F1 scores between \simlength and BLEU. Other parts-of-speech like symbols and interjections have high F1 scores as well, and words belonging to these classes are both relatively rare and highly discriminative regarding the semantics of the sentence.\footnote{Note that in the data, interjections (INTJ) often correspond to words like {\it Yes} and {\it No} which tend to be very important regarding the semantics of the translation in these cases.}
In contrast, parts-of-speech that in general convey little semantic information and are more common, like determiners, show very little difference in F1 between the two approaches.

\section{Qualitative Analysis}

\begin{table*}[ht!]
    \centering
    \small
    \begin{tabular}{| l | p{6.7cm} | c | c | c | c |}
    \hline
    System & Sentence & BLEU & SIM & $\Delta \text{BLEU}$ & $\Delta \text{SIM}$ \\
    \hline
    Reference & Workers have begun to clean up in R\"{o}szke. & - & - & - & - \\
    BLEU & Workers are beginning to clean up workers. &  29.15 & 69.12 & - & - \\
    \simlength & In R\"{o}szke, workers are beginning to clean up. & 25.97 & 95.39 & -3.18 & 26.27 \\
    \hline
    Reference & All that stuff sure does take a toll. & - & - & - & - \\
    BLEU & None of this takes a toll. & 25.98 & 54.52 & - & - \\
    \simlength & All of this is certain to take its toll. & 18.85 & 77.20 & -7.13 & 32.46 \\
    \hline
    \hline
    Reference & Another advantage is that they have fewer enemies. & - & - & - & - \\
    BLEU & Another benefit : they have less enemies. & 24.51 & 81.20 & - & - \\
    \simlength & Another advantage: they have fewer enemies. & 58.30 & 90.76 & 56.69 & 9.56 \\
    \hline
    Reference & I don't know how to explain - it's really unique. & - & - & - & - \\
    BLEU & I do not know how to explain it - it is really unique. & 39.13 & 97.42 & - & - \\
    \simlength & I don't know how to explain - it is really unique. & 78.25 & 99.57 & 39.12 & 2.15 \\
    \hline
    \end{tabular}
    \caption{Translation examples where the $|\Delta \text{BLEU}| - |\Delta \text{SIM}|$ statistic is among the highest and lowest in the validation set. The top two rows show examples where the generated sentences have similar sentence-level BLEU scores but quite different SIM scores. The bottom two rows show the converse. Negative values indicate the \simlength system had a higher score for that sentence. 
    }
    \label{table:metrics}
\end{table*}

We show examples of the output of all three systems in Table~\ref{table:examples} from the test sets, along with their human scores which are on a 0-5 scale. The first 5 examples show cases where \simlength better captures the semantics than BLEU or MLE. In the first three, the \simlength model adds a crucial word that the other two systems omit. This makes a significant difference in preserving the semantics of the translation. These words include verbs ({\it tells}), prepositions ({\it For}), adverbs ({\it viable}) and nouns ({\it conversation}). The fourth and fifth examples also show how \simlength can lead to more fluent outputs and is effective on longer sentences.

The last two examples are failure cases of using \simlength. In the first, it repeats a phrase, just as the MLE model does and is unable to smooth it out as the BLEU model is able to do. In the last example, \simlength again tries to include words ({\it Dr. Caglar}) significant  to the semantics of the sentence. However it misses on the rest of translation, despite being the only system to include this noun phrase.

\section{Metric Comparison}

We took all outputs of the validation set of the \texttt{de-en} data for our best \simlength and BLEU models, as measured by BLEU validation scores, and we sorted the outputs by the following statistic:

\begin{equation}
|\Delta \text{BLEU}| - |\Delta \text{SIM}|
\end{equation}
where BLEU in this case refers to sentence-level BLEU. Examples of some of the highest and lowest scoring sentence pairs are shown in Table~\ref{table:metrics} along with the system they came from (either trained with a BLEU cost or \simlength cost). 

The top half of the table shows examples where the difference in SIM scores is large, but the difference in BLEU scores is small. From these examples, we see that when SIM scores are very different, there is a difference in the meanings of the generated sentences. However, when the BLEU scores are very close, this is not the case. In fact, in these examples, less accurate translations have higher BLEU scores than more accurate ones. In the first sentence, an important clause is left out ({\it in R\"{o}szke}) and in the second, the generated sentence from the BLEU system actually negates the reference, despite having a higher BLEU score than the sentence from the \simlength system. 

Conversely, the bottom half of the table shows examples where the difference in BLEU scores is large, but the difference in SIM scores is small. From these examples, we can see that when BLEU scores are very different, the semantics of the sentence can still be preserved. However, the SIM score of these generated sentences with the references are close to each other, as we would hope to see. These examples illustrate a well-known problem with BLEU where synonyms, punctuation changes, and other small deviations from the reference can have a large impact on the score. As can be seen from the examples, these are less of a problem for the SIM metric.

\section{Related Work}
The seminal work on training machine translation systems to optimize particular evaluation measures was performed by \citet{P03-1021}, who introduced minimum error rate training (MERT) and used it to optimize several different metrics in statistical MT (SMT).
This was followed by a large number of alternative methods for optimizing machine translation systems based on minimum risk \cite{P06-2101}, maximum margin \cite{watanabe2007online}, or ranking \cite{hopkins2011tuning}, among many others.

Within the context of SMT, there have also been studies on the stability of particular metrics for optimization.
\citet{N10-1080} compared several metrics to optimize for SMT, finding BLEU to be robust as a training metric and finding that the most effective and most stable metrics for training are not necessarily the same as the best metrics for automatic evaluation. 
The WMT shared tasks included tunable metric tasks in 2011~\citep{W11-2103} and again in 2015~\citep{W15-3032} and 2016~\citep{W16-2303}. In these tasks, participants submitted metrics to optimize during training or combinations of metrics and optimizers, given a fixed SMT system. The 2011 results showed that nearly all metrics performed similarly to one another. The 2015 and 2016 results showed more variation among metrics, but also found that BLEU was a strong choice overall, echoing the results of \citet{N10-1080}. We have shown that our metric stabilizes training for NMT more than BLEU, which is a promising result given the limited success of the broad spectrum of previous attempts to discover easily tunable metrics in the context of SMT.

Some researchers have found success in terms of improved human judgments when training to maximize metrics other than BLEU for SMT. 
\citet{P13-2067} and \citet{beloucif2014improving} trained SMT systems to maximize variants of MEANT, a metric based on semantic roles.  \citet{liu-dahlmeier-ng:2011:EMNLP} trained systems using TESLA, a family of metrics based on softly matching $n$-grams using lemmas, WordNet synsets, and part-of-speech tags. 
We have demonstrated that our metric similarly leads to gains in performance as assessed by human annotators, and our method has an auxiliary advantage of being much simpler than these previous hand-engineered measures.

\citet{P16-1159} explored minimum risk training for NMT, 
finding that a
sentence-level BLEU score 
led to the best performance even when evaluated under other metrics. These results differ from the usual results obtained for SMT systems, in which tuning to optimize a metric leads to the best performance on that metric~\citep{P03-1021}. 
\citet{edunov2018classical} compared structured losses for NMT, also using  sentence-level BLEU.
They found risk to be an effective and robust choice, so we use risk as well in this paper.

\section{Conclusion}
We have proposed \simlength, an alternative to BLEU for use as a reward in minimum risk training. We have found that \simlength not only outperforms BLEU on automatic evaluations,
it correlates better with human judgments as well. Our analysis also shows that using this metric eases optimization and the translations tend to be richer in correct, semantically important words.
 
This is the first time to our knowledge that a continuous metric of semantic similarity
has been proposed for NMT optimization and shown to outperform sentence-level BLEU, and we hope that this can be the starting point for more research in this direction.

\appendix
\section{Appendix}
\subsection{Annotation Instructions} \label{appendix}
Below are the annotation instructions used by translators for evaluation.

\begin{itemize}
    \item 0. The meaning is completely different or the output is meaningless 
    \item 1. The topic is the same but the meaning is different
    \item 2. Some key information is different
    \item 3. The key information is the same but the details differ
    \item 4. Meaning is essentially equal but some expressions are unnatural
    \item 5. Meaning is essentially equal and the two sentences are well-formed English
\end{itemize}

\bibliography{acl2019}
\bibliographystyle{acl_natbib}

\end{document}